\documentclass[letterpaper, 10 pt, conference]{ieeeconf}  

\IEEEoverridecommandlockouts                              

\usepackage{float}
\usepackage{graphicx}
\usepackage{bm}
\usepackage{amsmath}
\usepackage{booktabs}
\usepackage{multirow}
\usepackage{algorithm}
\usepackage{algorithmic}

\title{\LARGE \bf
Stereo Event-based Visual-Inertial Odometry
}

\author{Kunfeng Wang, Kaichun Zhao, and Zheng You
\thanks{Kunfeng Wang, Kaichun Zhao and Zheng You (Corresponding author) are with the Department of Precision Instrument at the Tsinghua University, Beijing, China. (E-mail: wkf18@mails.tsinghua.edu.cn, kaichunz@tsinghua.edu.cn,yz-dpi@mail.tsinghua.edu.cn).}%
}

\begin{document}

\maketitle
\thispagestyle{empty}
\pagestyle{empty}

\begin{abstract}

Event-based cameras are new type vision sensors whose pixels work independently and respond asynchronously to brightness change with microsecond resolution, instead of providing standard intensity frames. Compared with traditional cameras, event-based cameras have low latency, no motion blur, and high dynamic range (HDR), which provide possibilities for robots to deal with some challenging scenes. We propose a visual-inertial odometry for stereo event-based cameras based on Error-State Kalman Filter (ESKF). The vision module updates the pose relies on the edge alignment of a semi-dense 3D map to a 2D image, and the IMU module updates pose by median integral. We evaluate our method on public datasets with general 6-DoF motion and compare the results against ground truth. We show that our proposed pipeline provides improved accuracy over the result of the state-of-the-art visual odometry  for stereo event-based cameras, while running in real-time on a standard CPU (low-resolution cameras). To the best of our knowledge, this is the first published visual-inertial odometry for stereo event-based cameras.

\end{abstract}

\section*{MULTIMEDIA MATARIAL}

Video: https://youtu.be/IclaeypKIPc

Code: https://github.com/WKunFeng/SEVIO

\section{INTRODUCTION}

Simultaneous localization and mapping (SLAM) has important applications in many emerging technologies such as robotics, intelligent transportation, and augmented/virtual reality (AR/VR). There are already a lot of works on slam based on traditional cameras [1], [2]. However, traditional cameras may fail in some challenging situations such as high-speed motions or high dynamic range scenes.

Event-based cameras (or event cameras) are bio-inspired vision sensors working very different from traditional cameras which report the pixel-wise intensity changes asynchronously at the time they occur, called “events” [3], [4], where each event consists of its spatio-temporal coordinates and the sign of the brightness change (e.g., 0 or 1). Event cameras have different types of sensors, such as Dynamic Vision Sensor (DVS) [3], DAVIS or ATIS. They don’t output an intensity image at a fixed rate but a stream of asynchronously events at microsecond resolution. Event cameras have numerous advantages over traditional cameras, such as microsecond latency, low power consumption, and a very high dynamic range (e.g., 140 dB compared to 60 dB of traditional cameras). With microsecond resolution, event cameras can work at high-speed motions, which will cause severe motion blur on traditional cameras. High dynamic range allow them to be used on broad illumination scenes. Thus, event cameras have the potential to tackle challenging scenarios in robotics.

\begin{figure}
\centering
\includegraphics[width=3.4 in]{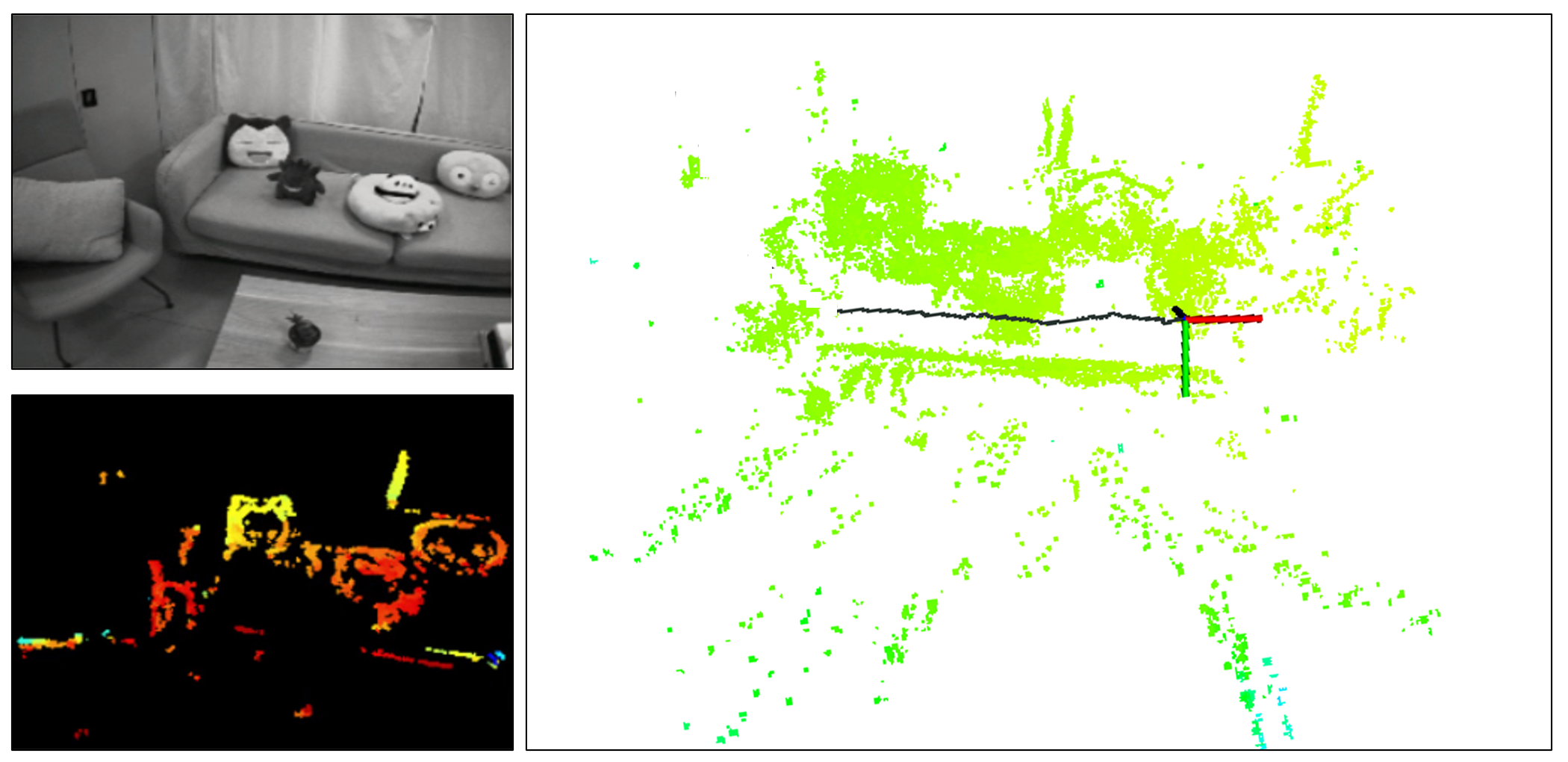}
\caption{Top left: scene. Bottom left: inverse depth map at time $t$. Right: global map and pose estimation. }
\label{fig1}
\end{figure}

The main challenge is how to unlock the potential of event cameras for SLAM and other vision tasks [4]. This is due to the fact that the output of event cameras is different from that of traditional cameras, so that frame-based vision algorithms designed for image sequences are not directly applicable to event data. Thus, we need new methods to process the data from these novel cameras. There are already some works for event cameras, such as feature tracking [5], [6], 3D orientation estimation [7], [8], [9], [10], and simultaneous localization and mapping (SLAM) [11], [12], [13], [14], [15].

However, vision-based SLAM/VO algorithms generally are less accurate and lack robustness in some challenging conditions, and fusion with additional sensors such as inertial measurement unit (IMU) is a common solution. In this paper we propose a stereo visual-inertial odometry (VIO) for event cameras with arbitrary 6-DoF motion (Fig.1). Our pipeline has three parts: vision module,IMU integral and ESKF. For the vision module, we reference the strategy of ESVO [15]. IMU data using the median integral, and fusion strategy is ESKF. Our contributions are as follows:

\begin{itemize}

\item A novel visual-inertial odometry for stereo event cameras. To the best of our knowledge, this is the first published visual-inertial odometry for stereo event cameras.
\item Our system does not rely on or produces traditional intensity images during computation, showing the potential of event data for SLAM/VO task.
\item A quantitative evaluation of our pipeline compared with ESVO on the public event camera datasets, demonstrating that our system is more accurate. The code has been open sourced.

\end{itemize}

\section{RELATED WORK}

In the past few years, many works have been proposed to use event cameras for ego-motion estimation. Here we review some of those literature [4].

\subsection{Event-based Depth Estimation}

a) monocular: Depth estimation with a single event camera has been shown in [11], [16], [18]. It is a significantly different problem from previous ones because it is difficult to match events across time. These methods produce a semi-dense 3D reconstruction (i.e., 3D edge map) of the scene with the information of the camera motion over time. Thus, they do not pursue instantaneous depth estimation, but rather depth estimation for SLAM/VO.

b) stereo: Most works on depth estimation with event cameras using events on a very short time (ideally on a per-event basis) from two or more synchronized cameras that are rigidly attached. The events from different camera image planes share a common clock. Then, following the classical two-step stereo paradigm, first match the events across image planes and then triangulate the location of the 3D point [19]. The problem is finding correspondences between events. Events are matched (i) using traditional stereo metrics (e.g., normalized cross-correlation) on event frames [20], [21] or time-surface [17], [22] or (ii) by exploiting simultaneity and temporal correlations of the events across sensors [22], [23], [24]. Most of these methods are demonstrated in scenes with static cameras and few moving objects, so that correspond-ences are easy to find.

\subsection{Event-based 3-DOF Estimation}

3-DOF estimation include rotation [7], [8], [9], [10] and planer motion [25]. 

Cook et al. [7] proposed an algorithm to jointly estimate ego-motion, image intensity and optical flow from events with an interacting network. However, it only applies to pure rotational motion. The filter based approach in [8] used probabilistic filters in parallel to track the 3D orientation of a rotating event camera and created high resolution panoramas of natural scenes. Rotational motion estimation was also presented in [9], where camera tracking was performed by minimization of a photometric error at the event locations given a probabilistic edge map. A motion compensation optimization framework was introduced in [10] to estimate the angular velocity of the camera rather than its absolute rotation. All of the above works are limited to rotation estimation, not translation.

An event-based 2D SLAM system was presented in [25], and it was restricted to planar motion and high contrast scenes. This method used a particle filter for tracking, with the event likelihood function inversely related to the reprojection error of the event with respect to the map. 

\subsection{Event-based VO}

Solutions to the problem of event-based 3D SLAM/VO for 6-DOF motions and natural scenes, with pure event cameras, have been proposed [11], [12], [13], [14], [15]. 

a) monocular: The approach in [11] extends [8] and consists of three interleaved probabilistic filters to perform pose tracking as well as depth and intensity estimation. However, it is computationally intensive, requiring a GPU for real-time operation. In contrast, the semi-dense approach in [12] shows that intensity reconstruction is not needed for depth estimation or pose tracking. The approach performs space sweeping for 3D reconstruction [16] and edge-map alignment (non-linear optimization with few events per frame) for pose tracking. The resulting SLAM system runs in real-time on a CPU. Formulation in [13] is underpinned by a principled joint optimisation problem involving non-parametric Gaussian Process motion modelling and incremental maximum a posteriori inference. However, its computational power consumption is also high, making it difficult to achieve real-time performance. [14] explores a new event-based line-SLAM approach following a parallel tracking and mapping philosophy.

b) stereo: The approach proposed in [15] tackle the problem of stereo visual odometry with event cameras in natural scenes and arbitrary 6-DoF motion in real time with a standard CPU. It reconstructs a semi-dense 3D map following two steps: (i) computing depth estimates of events and (ii) fusing such depth estimates into an accurate and populated depth map [17].

\subsection{Event-based VIO}

To improve the accuracy and robustness of visual odometry, combining with other sensors is a common option, such as IMU.

[26] tracked features using [5], and combined them with IMU data by a way of Kalman filter. [27] proposed to synthesize motion compensated event images from spatio-temporal windows of events and then detect-and-track features. Feature tracking were fused with inertial data by means of keyframe-based nonlinear optimization to recover the camera trajectory and a sparse map of 3D landmarks. As opposed to the above mentioned feature-based methods, the work in [28] optimizes a combined objective with inertial and event-reprojection error terms over a segment of the camera trajectory, in the style of visual-inertial bundle adjustment. [29] introduce an optimization-based framework using Lines.

There are also some works combine images, events and IMU data, such as [30], [31]. In this article, we want to show the characteristics of event cameras, so we do not provide a detailed introduction to these work.

\section{VISUAL-INERTIAL PIPELINE}

This section presents the details of stereo event-based visual-inertial odometry. We start with an overview of the entire pipeline. Then, we introduce how to process event data so that we can use its information efficiently. After that, we give the details of vision module and IMU integral module. At least, we present how to fuse information from different sensors with ESKF.

\subsection{Framework Overview}

\begin{figure*}
\centering
\includegraphics[width=6.7 in]{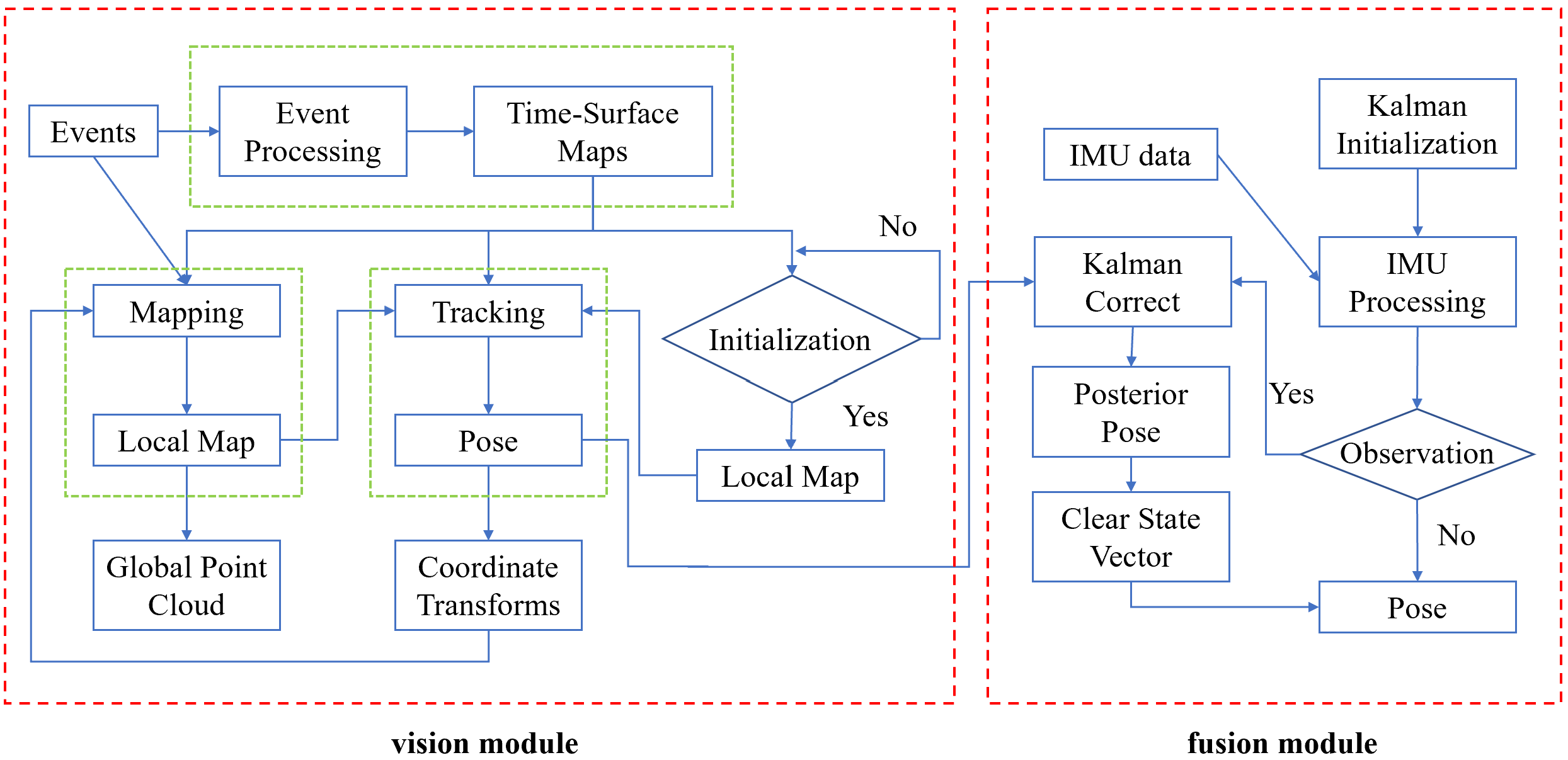}
\caption{Overview of our proposed stereo event-based visual-inertial odometry}
\label{fig2}
\end{figure*}

A flowchart of our proposed pipeline detailing all steps is illustrated in Figure 2. Let us briefly introduce the functionality of each module and explain how they work. First of all, the event processing module translate event stream to time-surface images, which will be used by vision module (Section \uppercase\expandafter{\romannumeral3}-B). Secondly, after an initialization phase, the tracking thread of vision module continuously estimates the pose of the left event camera by 3D-2D edge alignment. The mapping thread of vision module uses the events, time-surfaces and pose estimates to update a local map (semi-dense depth map), which is used by the tracking thread (Section \uppercase\expandafter{\romannumeral3}-C). At the same time, IMU module continues to estimate pose by integral. When the mapping thread is working, ESKF module receives an observation and performs a state update (Section \uppercase\expandafter{\romannumeral3}-D). Then, mapping thread uses the fused pose to refresh the local map.

Initialization: Vision module provides a coarse initial map by a stereo method (a modified SGM method [32]). The IMU module assumes that the initial state of the system is static [33], [34], using the first 1-2s (depending on dataset) of IMU data to estimate the gravity direction and the biases of the gyroscope and accelerometer.

\subsection{Event Representation (Time-Surface)}

\begin{figure}
\centering
\includegraphics[width=3.2 in]{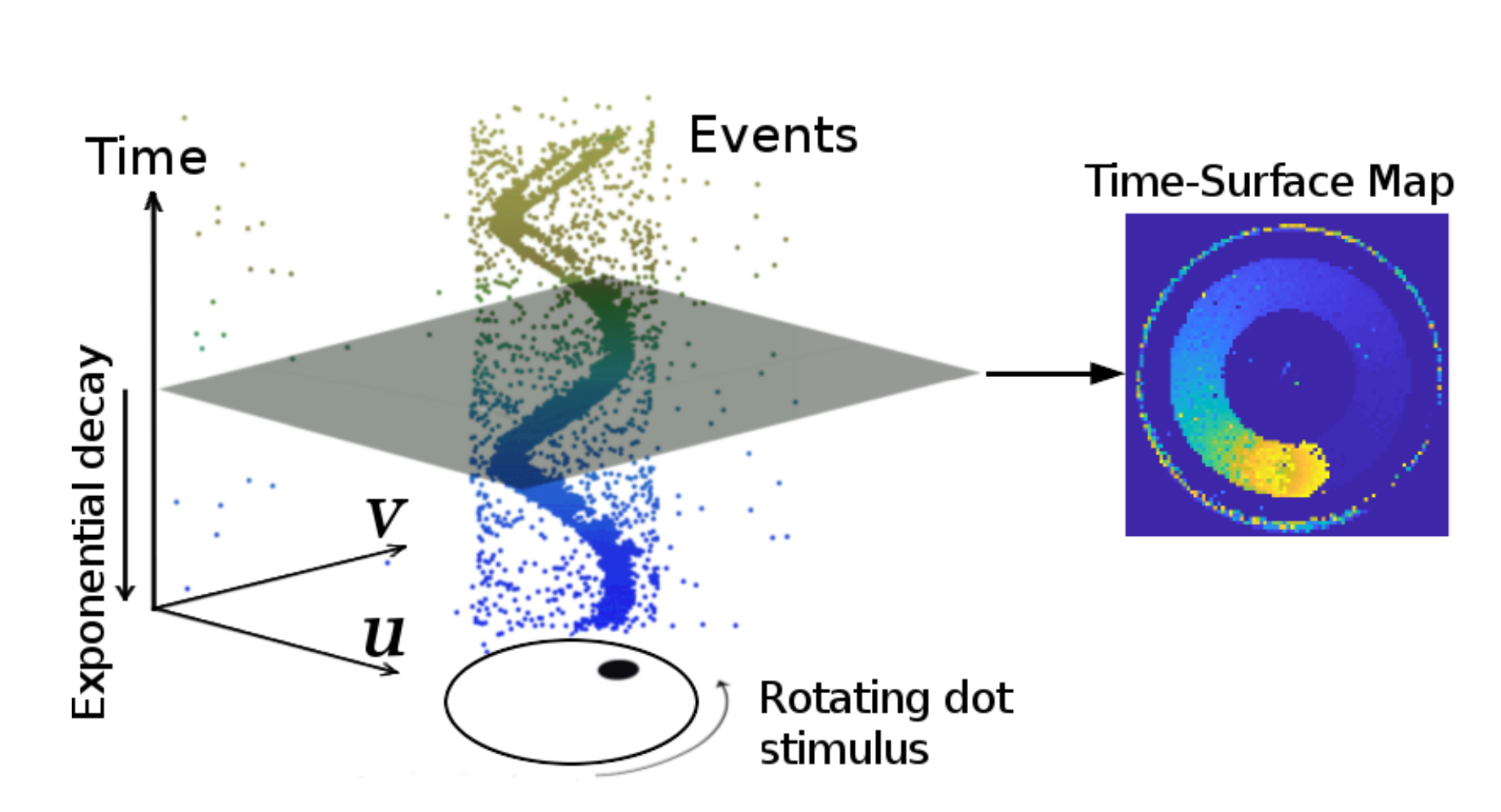}
\caption{Event Representation. Left: output of an event camera. Right: time-surface map. Figure adapted from [35]}
\label{fig3}
\end{figure}

The output of an event camera is a stream of asynchronous events. Each event can be described as
$ e_{k}=(u_k,v_k,t_k,p_k) $, consists of pixel coordinate $ (u_k,v_k)^{T} $, where intensity change of predefined size happened, timestamp $ t_k $, and polarity $ p_k (\{-1, 1\} \ or\  \{0, 1\}) $ of the change.

Generally, we do not process events asynchronously at the very high rate they occur, but use an alternative representation called Time-Surface (Fig.3). A time-surface (TS) is a 2D image where each pixel stores a single time value (e.g., the timestamp of the last event at that pixel [36], [37]), so that events are converted into an image whose “intensity” is a function of the motion history at that coordinate, and larger values means closer motion history. The value at each pixel coordinate 
$ \mathbf{x}=(u,v)^{T} $ is defined by:
\begin{equation}
\label{eq1}
\Gamma \doteq exp(- \frac{t-t_{last}(\mathbf{x})}{\eta})
\end{equation}
where $t$ is an arbitrary time $(t>t_{last}(\mathbf{x}))$, and $ t_{last}(\mathbf{x}) $ is the timestamp of the last event occurring at $ \mathbf{x} $. $ \eta $ denotes the constant decay rate parameter, which usually is small. The time-surface visualizes the history of moving brightness change at each pixel location, which usually presents the edges of the scenes. The pixel values in a time-surface are rescaled from [0, 1] to the range [0, 255] for convenient visualization and processing.

\subsection{Vision Module}

This moudle is mainly divided into two parts: mapping and tracking. Mapping thread estimates depth of each event and build a 3D map. Tracking thread updates the pose relies on 3D-2D edge alignment.


\subsubsection{mapping}

Assuming that we have a pair of time-surfaces ($\Gamma_{left}(\cdot,t),\Gamma_{right}(\cdot,t)$) at time $t$. The inverse depth $\rho^{*}$ of an event $(e_{t- \epsilon}= (\mathbf{x},t- \epsilon), \epsilon \in [0, \delta t])$ on the left image plane, which follows a camera trajectory  $ \mathbf{T}_{t-\delta t : t} $, is estimated by optimizing the objective function:

\begin{equation}
\label{eq2}
\rho^{*} = \mathop{argmin}\limits_{\rho} C (\mathbf{x},\rho,\Gamma_{left}(\cdot,t),\Gamma_{right}(\cdot,t),\mathbf{T}_{t-\delta t : t})
\end{equation}

\begin{equation}
\label{eq3}
C \doteq \sum\limits_{\mathbf{x}_{1,i}\in W_1,\mathbf{x}_{2,i}\in W_2} r_{i}^{2}(\rho)
\end{equation}

The residual:

\begin{equation}
\label{eq4}
r_{i}(\rho) \doteq \Gamma_{left}(\mathbf{x}_{1,i},t) - \Gamma_{right}(\mathbf{x}_{2,i},t)
\end{equation}
denotes the temporal difference between two corresponding pixels $ \mathbf{x}_{1,i} $ and $ \mathbf{x}_{2,i} $ inside neighborhoods (i.e., patches) $W_1$ and $W_2$, centered at $\mathbf{x}_1$ and $\mathbf{x}_2$ respectively. The points $\mathbf{x}_1$ and $\mathbf{x}_2$ are given by

\begin{equation}
\label{eq5}
\begin{aligned}
\mathbf{x}_1 & = \pi (^{c_t} \mathbf{T}_{c_{t- \epsilon}} \cdot \pi^{-1}(\mathbf{x},\rho_k)) \\
\mathbf{x}_2 & = \pi (^{right} \mathbf{T}_{left} \cdot ^{c_t} \mathbf{T}_{c_{t- \epsilon}} \cdot \pi^{-1}(\mathbf{x},\rho_k)) \\
\end{aligned}
\end{equation}

 The function $\pi$ projects a 3D point onto the camera's image plane, and $\pi^{-1}$ back-projects a pixel into 3D space with inverse depth $\rho$. $^{right} \mathbf{T}_{left}$ is the transformation
from the left to the right event camera. Suppose that the depth of each event point is known. Then, fusing inverse depth estimates to produce a semi-dense depth map at the current time, which is used for tracking later.

\subsubsection{tracking}

Time-surfaces record the motion history of the edges in the scene. Large values correspond to events that are close to the current time. To construct a minimum optimization problem, we use time-surface negative rather than time-surface, defining by

\begin{equation}
\label{eq6}
\overline{\Gamma}(\mathbf{x},t) = 1 - \Gamma(\mathbf{x},t)
\end{equation}

$\overline{\Gamma}(\mathbf{x},t)$ is also rescaled to the range $[0, 255]$. The tracking problem can be formulated as follows. Let $ \mathcal{S}^{\mathcal{F}_{ref}} = \{\mathbf{x}_i\}$ be a set of pixel locations with valid inverse depth $\rho_{i}$ in the reference frame $ \mathcal{F}_{ref}$. Assuming the TS negative at time k is available, denoted by $\overline{\Gamma}_{left}(\cdot,k)$, the goal is to find the pose $T$, which makes the warped semi-dense map $ T(\mathcal{S}^{\mathcal{F}_{ref}})$ aligns well with $\overline{\Gamma}_{left}(\cdot,k)$. This problem can be defined as

\begin{equation}
\label{eq7}
\bm{\psi}^{*} = \mathop{argmin}\limits_{\bm{\psi}} \sum\limits_{\mathbf{x} \in \mathcal{S}^{\mathcal{F}_{ref}}}(\overline{\Gamma}_{left}(W(\mathbf{x},\rho;\bm\psi),k))^2
\end{equation}
where the warping function

\begin{equation}
\label{eq8}
W(\mathbf{x},\rho;\bm\psi) \doteq \pi_{left}(T(\pi_{ref}^{-1}(\mathbf{x},\rho),G(\bm\psi)))
\end{equation}
transfers points from $\mathcal{F}_{ref}$ to the current frame. $\bm\psi$ is a $6 \times 1$ vector representing rotation and  translation. The fuction $G(\bm\psi)$ provides the transformation matrix corresponding to $\bm\psi$. The function $\pi_{ref}^{-1}(\cdot)$ back-projects a pixel $\mathbf{x}$ into space with the known inverse depth $\rho$, while $\pi_{left}(\cdot)$ projects the transformed space point onto the image plane of the left camera. $T(\cdot)$ performs a change of coordinates,
transforming the 3D point with motion $G(\bm\psi)$ from $\mathcal{F}_{ref}$ to the left frame $\mathcal{F}_{k}$ of the current stereo observation.

\subsection{ESKF Description}

\subsubsection{Structure of the ESKF state vector}

The goal of the proposed filter is to track the pose of the IMU frame $\{\mathbf{I}\}$ (generally considered as body frame) with respect to a global frame of reference $\{\mathbf{G}\}$. An overview of the algorithm is given in Algorithm 1. The IMU data is processed for propagating the ESKF state and covariance. Then, each time an observation is received, updating the state vector. The IMU state is a $15 \times 1$ vector which is defined as:

\begin{equation}
\label{eq9}
\mathbf{x}_{IMU} = (\mathbf{p}^T,\mathbf{v}^T,\mathbf{q}^T,\mathbf{b}_a^T,\mathbf{b}_\omega^T)^T
\end{equation}
where the quaternion $\mathbf{q}$ represents the rotation from the frame $\{\mathbf{I}\}$ to the frame $\{\mathbf{G}\}$. The vectors $\mathbf{v} \in \mathbf{R}^3$ and $\mathbf{p} \in \mathbf{R}^3$ represent the velocity and position of the body frame $\{\mathbf{I}\}$ in the global frame $\{\mathbf{G}\}$. The vectors $\mathbf{b}_a \in \mathbf{R}^3$ and $\mathbf{b}_\omega \in \mathbf{R}^3$ are the biases of the linear acceleration and angular velocity from the IMU, which are modeled as random walk processes, driven by the white Gaussian noise vectors $\mathbf{n}_{ba}$ and $\mathbf{n}_{b\omega}$ , respectively. Following (9), the IMU error-state is defined as:
\begin{equation}
\label{eq10}
\delta{\mathbf{x}_{IMU}} = (\delta{\mathbf{p}^T},\delta{\mathbf{v}^T},\delta{\mathbf{q}^T},\delta{\mathbf{b}_a^T},\delta{\mathbf{b}_\omega^T})^T
\end{equation}
the standard error definition is used for the position, velocity, and biases (e.g., $\widetilde{\mathbf{p}} = \mathbf{p} + \delta{\mathbf{p}}$, $\widetilde{\mathbf{p}}$ is the real value of position and $\mathbf{p}$ is ideal value). For the quaternion, a different error definition is employed, which is defined by the relation $\widetilde{\mathbf{q}} = \mathbf{q} \bigotimes \delta{\mathbf{q}}$. The symbol $\bigotimes$ denotes quaternion multiplication. The error quaternion is:
\begin{equation}
\label{eq11}
\delta{\mathbf{q}} = 
\begin{bmatrix}
1\\
\dfrac{\delta{\bm{\theta}}}{2}\\
\end{bmatrix}
\end{equation}
where $\delta{\bm{\theta}} $ represents a small angle rotation.

\subsubsection{Process model}

The continuous dynamics of the IMU error-state is:
\begin{equation}
\label{eq12}
\begin{aligned}
\delta{\dot{\mathbf{p}}} & = \delta{\mathbf{v}} \\
\delta{\dot{\mathbf{v}}} & = -\mathbf{R}_t [\mathbf{a}_t - \mathbf{b}_{a_t}]_{\times} \delta{\bm{\theta}} + \mathbf{R}_t(\mathbf{n}_a-\delta{\mathbf{b}_a})\\
\delta{\dot{\bm{\theta}}} & = -[\bm{\omega}_t-\mathbf{b}_{\omega_{t}}]_{\times} \delta{\bm{\theta}} + \mathbf{n}_{\omega} - \delta{\mathbf{b}}_{\omega} \\
\delta{\dot{\mathbf{b}_{a}}} & = \mathbf{n}_{ba} \\
\delta{\dot{\mathbf{b}_{\omega}}} & = \mathbf{n}_{b\omega}
\end{aligned}
\end{equation}

In these expressions, $\mathbf{a}_t$ and $\bm{\omega}_t$ are acceleration and angular velocity from IMU measurements. $[\cdot]_{\times}$ means skew symmetric matrix. $\mathbf{R}_{t}$ is the rotation matrix from frame $\{\mathbf{I}\}$ to frame $\{\mathbf{G}\}$, $\mathbf{n}_{a}$ and ${\mathbf{n}_{\omega}}$ are zero-mean, white Gaussian noise processes modeling the measurement noise. The linearized continuous-time model for the IMU error-state is:

\begin{equation}
\label{eq13}
\delta{\dot{\mathbf{x}}} = \mathbf{F}_{t} \delta{\mathbf{x}+\mathbf{B}_t \mathbf{n}}
\end{equation}
where $\mathbf{n} = (\mathbf{n}_{a}^{T},\mathbf{n}_{\omega}^{T},\mathbf{n}_{b_a}^{T},\mathbf{n}_{b_{\omega}}^{T})^T$ . The vectors $\mathbf{n}_{ba}$ and $\mathbf{n}_{b\omega}$ are the random walk rate of the accelerometer and gyroscope measurement biases. $\mathbf{F}_{t}$ and $\mathbf{B}_t$ are shown in Appendix A.

To deal with discrete time measurement from the IMU, we apply median integral to propagate the estimated IMU state: 
\begin{equation}
\label{eq14}
\begin{aligned}
\mathbf{q}_{\omega b_k} & = \mathbf{q}_{\omega b_{k-1}} \otimes 
\begin{bmatrix}
\cos \dfrac{\phi}{2}\\
\dfrac{\bm{\phi}}{\phi} \sin \dfrac{\phi}{2} \\
\end{bmatrix}\\
\mathbf{v}_k & = \mathbf{v}_{k-1} + (\dfrac{\mathbf{R}_{\omega b_k} \mathbf{a}_k + \mathbf{R}_{\omega b_{k-1}} \mathbf{a}_{k-1}}{2}-\mathbf{g})(t_k-t_{k-1})\\
\mathbf{p}_k & = \mathbf{p}_{k-1} + \dfrac{\mathbf{v}_k+\mathbf{v}_{k-1}}{2} (t_k-t_{k-1})
\end{aligned}
\end{equation}
where $ \bm{\phi} = \dfrac{\bm{\omega}_{k-1}+\bm{\omega}_{k}}{2} (t_k-t_{k-1})$, $ \phi = \Vert \bm{\phi} \Vert$.\\[2pt]

To discretize (13), 
\begin{equation}
\label{eq15}
\delta{\mathbf{x}}_k = \mathbf{F}_{k-1} \delta{\mathbf{x}}_{k-1}+\mathbf{B}_{k-1} \mathbf{n}_k
\end{equation}
where $\mathbf{F}_{k-1}$ and $\mathbf{B}_{k-1}$ are shown in Appendix B.

\subsubsection{Measurement Model}

Now we introduce the measurement model employed for updating the state estimates. Generally, the observation equation is written as:
\begin{equation}
\label{eq16}
\mathbf{y}=\mathbf{G}_t\delta\mathbf{x}+\mathbf{C}_t\mathbf{n}_R
\end{equation}

In this expression, $\mathbf{G}_t$ is the measurement Jacobian matrix, and $\mathbf{n}_R$ is observation noise.

\begin{equation}
\label{eq17}
\mathbf{n}_R = (n_{\delta{\overline{p}}_x},n_{\delta{\overline{p}}_y},n_{\delta{\overline{p}}_z},n_{\delta{\overline{\theta}}_x},n_{\delta{\overline{\theta}}_y},n_{\delta{\overline{\theta}}_z})^T
\end{equation}

The noise term $\mathbf{C}_t \mathbf{n}_R$ is zero-mean, white, and uncorrelated to the state, for the ESKF framework to be applied. In this application condition, the observation equation is:

\begin{equation}
\label{eq18}
\mathbf{y} = [\delta \overline{\mathbf{p}}^T , \delta \overline{\bm{\theta}}^T]^T
\end{equation}




According to the previous formulas, we get the equations of discrete ESKF, which are shown in Appendix C.

\begin{algorithm}
\caption{ Framework of SEVIO}   
\label{alg:Framwork}   
\begin{algorithmic}[1] 
\REQUIRE ~~\\ 
The event data from two event-based cameras;\\    
The acceleration and angular velocity from IMU.
\ENSURE ~~\\ 
The pose of the body frame $\{ \mathbf{I} \}$ with respect to the global frame $\{ \mathbf{G} \}$. 
\STATE Initialize: A modified SGM method (vision); Estimation of IMU biases and the gravity direction (ESKF).
\label{ code:fram:1 }
\STATE IMU propagation;   
\label{code:fram:2}
\IF{no observation} 
\STATE Posterior equals prior (Eq.20)
\ELSE
\STATE Update posterior pose (Eq.21)
\STATE  Clear the error-state(Eq.22)
\ENDIF 
\end{algorithmic}  
\end{algorithm}  

\subsubsection{Filter Update}

In the preceding sections, we present the process model and measurement model. Now we introduce in detail the updating phase of the ESKF. The whole process is summarized in Algorithm 1.

Before we start the ESKF process, we need to know the initial state of the system. In our implementation, we assume that the initial state of the system is static and take the average value of IMU output for a period of time to estimate the gravity direction and the biases of the gyroscope and accelerometer. Then, we get the rotation from the IMU frame to the world. Next, initializing filter parameters, including state vector $\delta \hat{\mathbf{x}}_0$, variance $\hat{\mathbf{P}}_0$, process noise $\mathbf{Q}$ and observation noise $\mathbf{R}$. Specific form is shown in Appendix D.

Then, we integrate the IMU output according to (14) for performing filter prior estimate and executing the first two steps of ESKF's five steps:

\begin{equation}
\label{eq21}
\begin{aligned}
\delta \check{\mathbf{x}}_k & = \mathbf{F}_{k-1} \delta \hat{\mathbf{x}}_{k-1} + \mathbf{B}_{k-1} \mathbf{n}_{k}\\
\check{\mathbf{P}}_k & = \mathbf{F}_{k-1} \hat{\mathbf{P}}_{k-1} \mathbf{F}_{k-1}^T + \mathbf{B}_{k-1} \mathbf{Q}_k \mathbf{B}_{k-1}^T \\
\end{aligned}
\end{equation}

When there is no observation, posterior equals prior:
\begin{equation}
\label{eq22}
\begin{aligned}
\delta \hat{\mathbf{x}}_k & = \delta \check{\mathbf{x}}_k \\
\hat{\mathbf{P}}_k & = \check{\mathbf{P}}_k \\
\hat{\mathbf{x}}_k & = \check{\mathbf{x}}_k \\
\end{aligned}
\end{equation}

When the mapping thread of vision module works, we consider an observation is received and execute the last three steps of ESKF's five steps:

\begin{equation}
\label{eq23}
\begin{aligned}
\mathbf{K}_k & = \check{\mathbf{P}}_k \mathbf{G}_{k}^{T} (\mathbf{G}_k \check{\mathbf{P}}_k \mathbf{G}_k^T + \mathbf{C}_k \mathbf{R}_k \mathbf{C}_k^T)^{-1}\\
\hat{\mathbf{P}}_k & = (\mathbf{I}-\mathbf{K}_k \mathbf{G}_k) \check{\mathbf{P}}_k \\
\delta \hat{\mathbf{x}}_k & = \delta \check{\mathbf{x}}_k + \mathbf{K}_k (\mathbf{y}_k - \mathbf{G}_k \delta \check{\mathbf{x}}_k) \\
\end{aligned}
\end{equation}

After that, we update the posterior pose and clear the state vector $\delta \hat{\mathbf{x}}_k$:

\begin{equation}
\label{eq24}
\begin{aligned}
\hat{\mathbf{p}}_k & = \check{\mathbf{p}}_k - \delta \hat{\mathbf{p}}_k \\
\hat{\mathbf{v}}_k & = \check{\mathbf{v}}_k - \delta \hat{\mathbf{v}}_k \\
\hat{\mathbf{R}}_k & = \check{\mathbf{R}}_k(\mathbf{I}-[\delta \hat{\bm{\theta}}_k]_{\times})\\
\hat{\mathbf{b}}_{ak} & = \check{\mathbf{b}}_{ak} - \delta \hat{\mathbf{b}}_{ak}\\
\hat{\mathbf{b}}_{\omega k} & = \check{\mathbf{b}}_{\omega k} - \delta \hat{\mathbf{b}}_{\omega k}\\
\end{aligned}
\end{equation}

\begin{figure*}
\centering
\includegraphics[width=6.6 in]{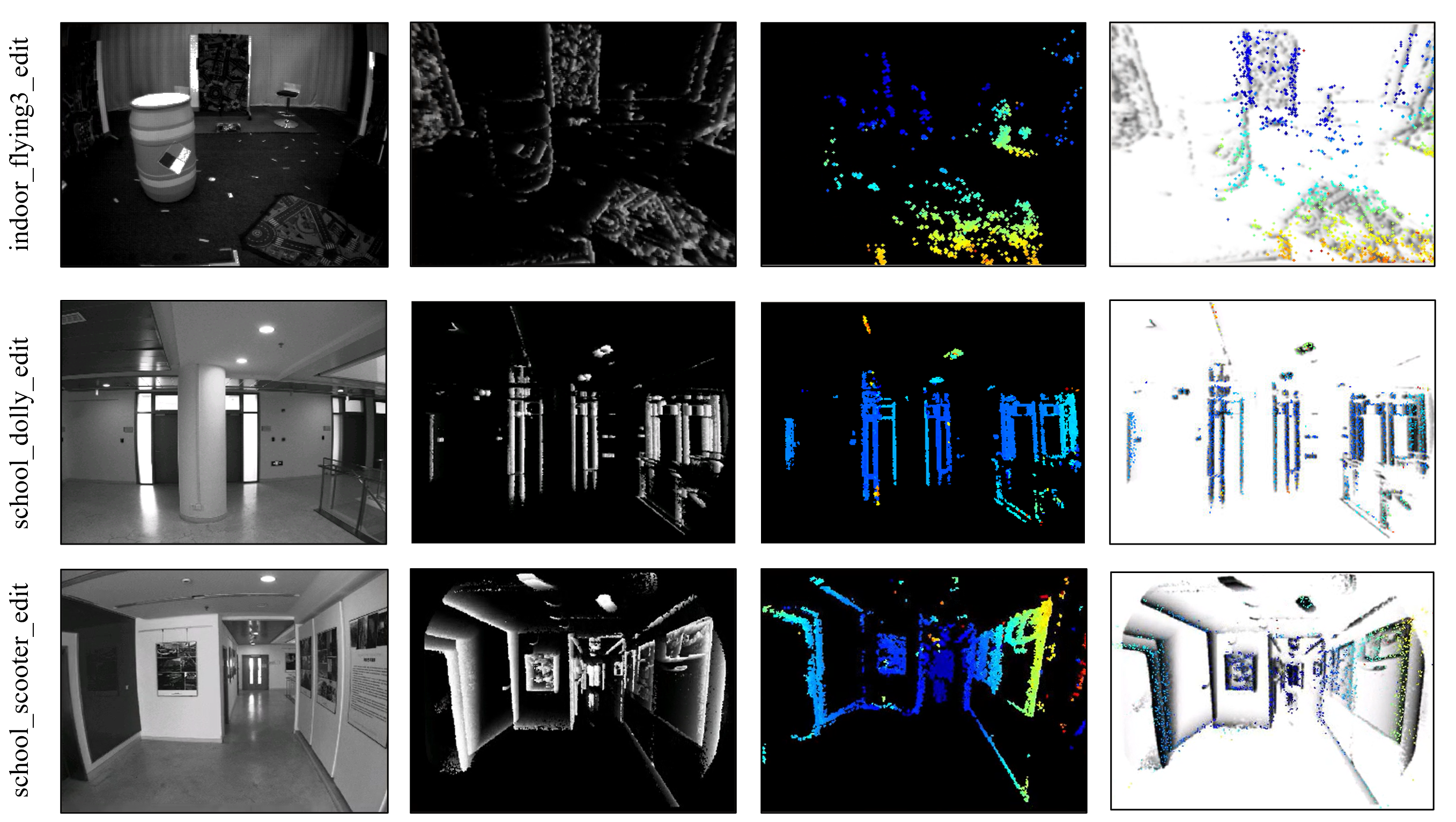}
\caption{The first column shows images from traditional camera. The second column is time-surface. The third column is inverse depth map. The last column is warping depth map overlaid on the time-surface negative.}
\label{fig5}
\end{figure*}

\section{EXPERIMENTS}

In this section, we present the datasets that we used and evaluate the proposed stereo event-based VIO system. The results show that SEVIO produces more accurate trajectories than ESVO. Some results are shown in Fig.4. Then, we provide a test in high dynamic environment to show the potential of event cameras. Finally, We illustrate the real-time performance of our pipeline on different resolution event data.

\begin{table}[H]
\caption{Parameters of stereo event-camera rigs used in the datasets.}
\label{tab1}
\centering
\begin{tabular}{c c c c}
\hline
Dataset & camera & resolution(pixel) & baseline(cm) \\
\hline
MVSEC & DAVIS 346 & $346 \times 260$ & 10.0 \\
VECtor & Prophesee Gen3 & $640 \times 480$ & 17.0 \\
\hline
\end{tabular}
\end{table}

\subsection{Datasets Used}

To evaluate the proposed stereo VIO system we use sequences from publicly available datasets [38], [39]. Dataset provided by [38] was collected from a handheld rig, a flying hexacopter, a car, and a motorcycle, with calibrated sensors data from different environments. Dataset [39] was collected using a simple handle for handheld, a wheeled tripod, and a helmet in an indoor environment.

Our algorithm works on undistorted and stereo-rectified coordinates. Cameras and imu calibration parameters are known in advance. We used partial sequences of the above dataset to verify our algorithm. The parameters of the stereo event cameras in each dataset used are listed in Table \uppercase\expandafter{\romannumeral1}.

\begin{table}
\caption{Absolute Pose Error and Relative Pose Error [RMSE(m)]}
\label{tab2}
\centering
\begin{tabular}{c c c c c}
\toprule
\multirow{2}{*}{Sequences} & \multicolumn{2}{c}{ESVO} & \multicolumn{2}{c}{SEVIO}\\
\cmidrule{2-5}
& APE & RPE & APE &RPE \\
\midrule
indoor\_flying1\_edit & \textbf{0.190} & 0.014 & 0.299 & \textbf{0.011} \\
indoor\_flying3\_edit & 0.342 & 0.027 & \textbf{0.266} & \textbf{0.010} \\
school\_dolly\_edit & 0.990 & 0.077 & \textbf{0.703} & \textbf{0.075} \\
school\_scooter\_edit & 2.666 & 0.233 & \textbf{1.291} & \textbf{0.195} \\
units\_dolly\_edit & 0.714 & 0.096 & \textbf{0.514} & \textbf{0.084} \\
\bottomrule
\end{tabular}
\end{table}

\subsection{Accuracy Evaluation}

To show the performance of the full VIO system, we report pose estimation results by common metrics: relative pose error (RPE) and absolute pose error (APE). Since no open-source event-based VIO projects is yet available, we only compared with the stereo visual pipeline (ESVO) with the same vision parameters. Due to the vision module is unstable so it is difficult to maintain long time working. To ensure that the trajectory is effective, we select the first part of the sequences for calculation and analysis. The sequences we used is also provided.

\begin{figure}
\centering
\includegraphics[width=3.3 in]{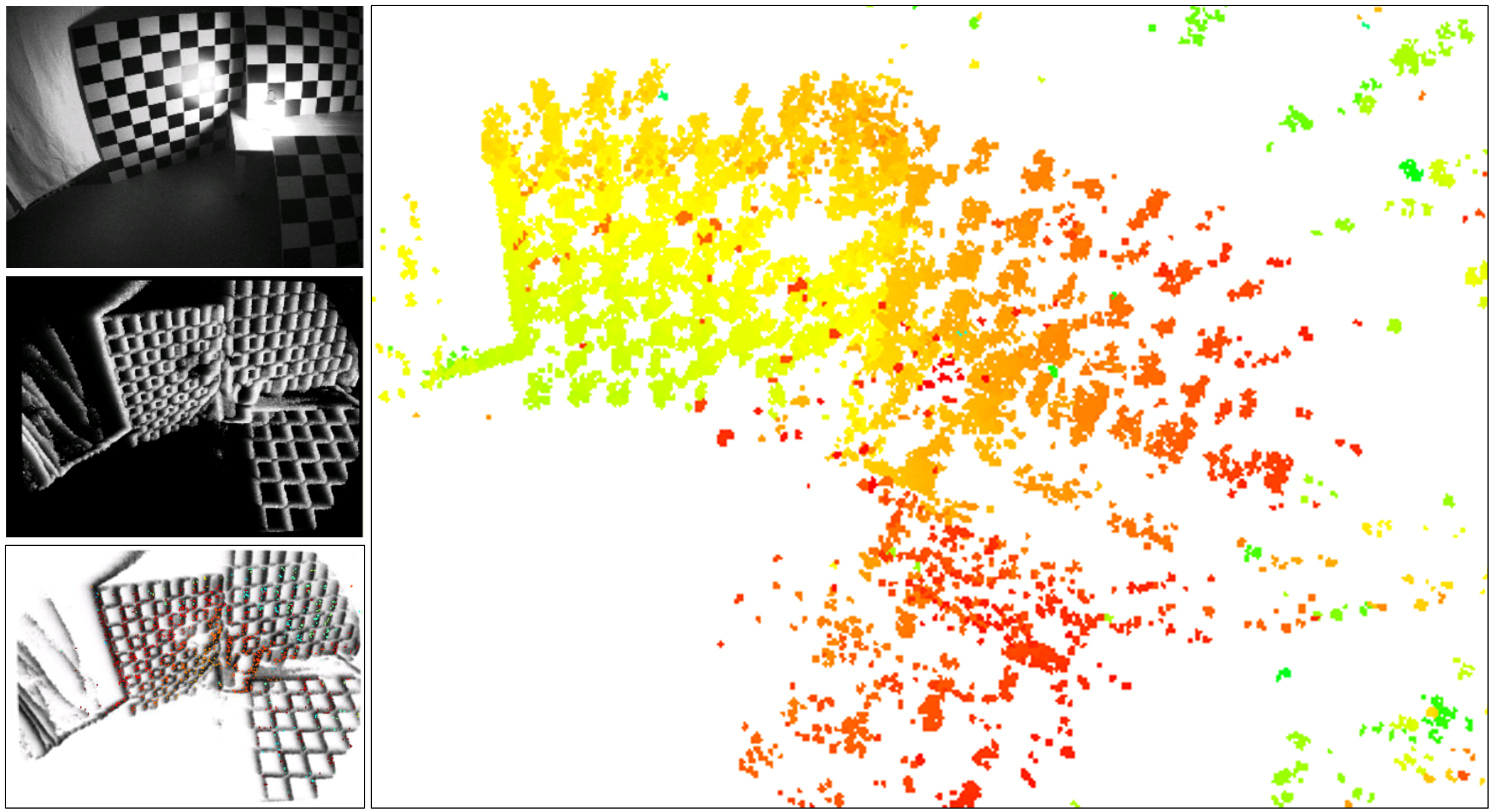}
\caption{The left column contains three images, which are the HDR scene, time-surface, and depth map  overlaid on the time-surface negative respectively. Right is the local map.}
\label{fig5}
\end{figure}

The evaluation is performed on five sequences with ground truth and the results are shown in Table \uppercase\expandafter{\romannumeral2}. The better results per sequence are highlighted in bold. It shows that most of the trajectory accuracy has been improved. Our method is less accurate than ESVO on indoor\_flying1\_edit. This is because after the initialization of IMU, vision still takes a period of time to initialize, resulting in obvious drift of IMU parameters. In addition, better vision and filter parameters may further improve the accuracy of the results.

\subsection{HDR Environment}
Event cameras are expected to solve some scenes where traditional cameras can not work and one of the advantages of event cameras is that they can work in HDR scenes. Event cameras have high dynamic range (with about 140 dB) campared with tradition cameras (with about 60dB). VECtor benchmark provide a sequence that have HDR scene. It is collected in a dark room with a lamp to increase the range of scene brightness variations. Under such conditions, the image quality of traditional cameras is seriously affected, which would cause the traditional algorithms to fail. As shown in Fig.5, event cameras still work well and build a 3D map by our pipeline.

\subsection{Real-time Performance}

We test the real-time performance of the algorithm on a standard CPU (Intel Xeon Platinum 8375C CPU @ 2.90GHz). Our system can work in real-time with event data from low resolution cameras (e.g., $346 \times 260$). For higher resolution cameras (e.g., $640 \times 480$), we need to reduce the play speed of the data for better performance (e.g., 0.2). Although event cameras only records edge information, the high frequency causes a large amount of data, resulting in significant computational power consumption. Choosing partly appropriate event data to participate in calculations may be a solution.

\section{CONCLUSION}

In this paper, we presented a stereo event-based visual-inertial pipeline using a ESKF framework. To the best of our knowledge, this is the first published work that solve this problem. Compared with the visual odometry for stereo event cameras (ESVO [15]), our method has higher accuracy. We also demonstrate the potential of event cameras in high dynamic environments. The algorithm and sequences used for evaluation have been open sourced. In the future work, the result with higher accuracy may be obtained by using fusion strategy based on optimization. Considering a solution to reduce the computational consumption of event camera data is also meaningful.

\section*{APPENDIX}

\subsection{The $\mathbf{F}_t$ and $\mathbf{B}_t$ in (13)}

\begin{equation}
\label{ap a1}
\mathbf{F}_t =
\begin{bmatrix}
\mathbf{0}_{3 \times 3} & \mathbf{I}_3 & \mathbf{0}_{3 \times 3} & \mathbf{0}_{3 \times 3} & \mathbf{0}_{3 \times 3}\\
\mathbf{0}_{3 \times 3} & \mathbf{0}_{3 \times 3} & -\mathbf{R}_t{} [\mathbf{a}_t -\mathbf{b}_{a_t}]_{\times} & -\mathbf{R}_t{} & \mathbf{0}_{3 \times 3}\\
\mathbf{0}_{3 \times 3} & \mathbf{0}_{3 \times 3} & -[\bm{\omega}_t - \mathbf{b}_{\omega_t}]_{\times} & \mathbf{0}_{3 \times 3} & -\mathbf{I}_3 \\
\mathbf{0}_{3 \times 3} & \mathbf{0}_{3 \times 3} & \mathbf{0}_{3 \times 3} & \mathbf{0}_{3 \times 3} & \mathbf{0}_{3 \times 3}\\
\mathbf{0}_{3 \times 3} & \mathbf{0}_{3 \times 3} & \mathbf{0}_{3 \times 3} & \mathbf{0}_{3 \times 3} & \mathbf{0}_{3 \times 3}\\
\end{bmatrix} \notag
\end{equation}

\begin{equation}
\label{ap a2}
\mathbf{B}_t =
\begin{bmatrix}
\mathbf{0}_{3 \times 3} & \mathbf{0}_{3 \times 3} & \mathbf{0}_{3 \times 3} & \mathbf{0}_{3 \times 3} \\
\mathbf{R}_t & \mathbf{0}_{3 \times 3} & \mathbf{0}_{3 \times 3} & \mathbf{0}_{3 \times 3} \\
\mathbf{0}_{3 \times 3} & \mathbf{I}_3 & \mathbf{0}_{3 \times 3} & \mathbf{0}_{3 \times 3} \\
\mathbf{0}_{3 \times 3} & \mathbf{0}_{3 \times 3} & \mathbf{I}_3 & \mathbf{0}_{3 \times 3} \\
\mathbf{0}_{3 \times 3} & \mathbf{0}_{3 \times 3} & \mathbf{0}_{3 \times 3} & \mathbf{I}_3 \\
\end{bmatrix} \notag
\end{equation}

\subsection{The $\mathbf{F}_{k-1}$ and $\mathbf{B}_{k-1}$ in (15)}

\begin{equation}
\label{ap b1}
\mathbf{F}_{k-1} = \mathbf{I}_{15}+\mathbf{F}_t T \notag
\end{equation}

\begin{equation}
\label{ap b2}
\mathbf{B}_{k-1} =
\begin{bmatrix}
\mathbf{0}_{3 \times 3} & \mathbf{0}_{3 \times 3} & \mathbf{0}_{3 \times 3} & \mathbf{0}_{3 \times 3} \\
\mathbf{R}_{k-1}T & \mathbf{0}_{3 \times 3} & \mathbf{0}_{3 \times 3} & \mathbf{0}_{3 \times 3} \\
\mathbf{0}_{3 \times 3} & \mathbf{I}_3 T & \mathbf{0}_{3 \times 3} & \mathbf{0}_{3 \times 3} \\
\mathbf{0}_{3 \times 3} & \mathbf{0}_{3 \times 3} & \mathbf{I}_3 \sqrt{T} & \mathbf{0}_{3 \times 3} \\
\mathbf{0}_{3 \times 3} & \mathbf{0}_{3 \times 3} & \mathbf{0}_{3 \times 3} & \mathbf{I}_3 \sqrt{T}\\
\end{bmatrix} \notag
\end{equation}

T is the period of the ESKF.

\subsection{Discrete ESKF equations}

\begin{equation}
\label{ap c1}
\begin{aligned}
\delta \check{\mathbf{x}}_k & = \mathbf{F}_{k-1} \delta \hat{\mathbf{x}}_{k-1} + \mathbf{B}_{k-1} \mathbf{n}_{k}\\
\check{\mathbf{P}}_k & = \mathbf{F}_{k-1} \hat{\mathbf{P}}_{k-1} \mathbf{F}_{k-1}^T + \mathbf{B}_{k-1} \mathbf{Q}_k \mathbf{B}_{k-1}^T \\
\mathbf{K}_k & = \check{\mathbf{P}}_k \mathbf{G}_{k}^{T} (\mathbf{G}_k \check{\mathbf{P}}_k \mathbf{G}_k^T + \mathbf{C}_k \mathbf{R}_k \mathbf{C}_k^T)^{-1}\\
\hat{\mathbf{P}}_k & = (\mathbf{I}-\mathbf{K}_k \mathbf{G}_k) \check{\mathbf{P}}_k \\
\delta \hat{\mathbf{x}}_k & = \delta \check{\mathbf{x}}_k + \mathbf{K}_k (\mathbf{y}_k - \mathbf{G}_k \delta \check{\mathbf{x}}_k) \\
\end{aligned} \notag
\end{equation}

\subsection{Variance, Process noise and observation noise}



Variance:

\begin{equation}
\label{ap d2}
\hat{\mathbf{P}}_0 =
\begin{bmatrix}
\mathbf{P}_{\delta p} & \mathbf{0}_{3 \times 3} & \mathbf{0}_{3 \times 3} & \mathbf{0}_{3 \times 3} & \mathbf{0}_{3 \times 3}\\
\mathbf{0}_{3 \times 3} & \mathbf{P}_{\delta v} & \mathbf{0}_{3 \times 3} & \mathbf{0}_{3 \times 3} & \mathbf{0}_{3 \times 3}\\
\mathbf{0}_{3 \times 3} & \mathbf{0}_{3 \times 3} & \mathbf{P}_{\delta \theta} & \mathbf{0}_{3 \times 3} & \mathbf{0}_{3 \times 3}\\
\mathbf{0}_{3 \times 3} & \mathbf{0}_{3 \times 3} & \mathbf{0}_{3 \times 3} & \mathbf{P}_{\delta b_a} & \mathbf{0}_{3 \times 3}\\
\mathbf{0}_{3 \times 3} & \mathbf{0}_{3 \times 3} & \mathbf{0}_{3 \times 3} & \mathbf{0}_{3 \times 3} & \mathbf{P}_{\delta b_{\omega}}\\
\end{bmatrix} \notag
\end{equation}

Process noise and observation noise:

\begin{equation}
\label{ap d3}
\mathbf{Q} =
\begin{bmatrix}
\mathbf{Q}_{a} & \mathbf{0}_{3 \times 3} & \mathbf{0}_{3 \times 3} & \mathbf{0}_{3 \times 3}\\
\mathbf{0}_{3 \times 3} & \mathbf{Q}_{\omega} & \mathbf{0}_{3 \times 3} & \mathbf{0}_{3 \times 3}\\
\mathbf{0}_{3 \times 3} & \mathbf{0}_{3 \times 3} & \mathbf{Q}_{b_a} & \mathbf{0}_{3 \times 3}\\
\mathbf{0}_{3 \times 3} & \mathbf{0}_{3 \times 3} & \mathbf{0}_{3 \times 3}& \mathbf{Q}_{b_{\omega}}\\
\end{bmatrix} \notag
\end{equation}

\begin{equation}
\label{ap d4}
\mathbf{R} =
\begin{bmatrix}
\mathbf{R}_{\delta p} & \mathbf{0}_{3 \times 3} \\
\mathbf{0}_{3 \times 3} & \mathbf{R}_{\delta \theta} \\
\end{bmatrix} \notag
\end{equation}

\end{document}